\pdfoutput=1

\documentclass[11pt]{article}
\usepackage{graphicx}
\graphicspath{{figs/}}
\usepackage[preprint]{acl}

\usepackage{times}
\usepackage{latexsym}

\usepackage[T1]{fontenc}

\usepackage[utf8]{inputenc}

\usepackage{microtype}
\usepackage{balance}
\usepackage{inconsolata}

\usepackage{multirow}
\usepackage{subfigure}
\usepackage{amsmath}
\usepackage{booktabs}
\usepackage{amssymb}

\def\model{GraphEdit}

\title{GraphEdit: Large Language Models for Graph Structure Learning}

\author{Zirui Guo$^{1,2}$, Lianghao Xia$^{2}$, Yanhua Yu$^{1,*}$, Yuling Wang$^{1}$,
\\ \textbf{Kangkang Lu}$^{1}$, \textbf{Zhiyong Huang}$^{3}$, \textbf{Chao Huang}$^{2}$ \\
Beijing University of Posts and Telecommunications$^1$, \\
University of Hong Kong$^2$, \\
National University of Singapore$^3$ \\
\{zrguo, yuyanhua\}@bupt.edu.cn, chaohuang75@gmail.com
}

\begin{document}
\maketitle
\begin{abstract}
Graph Structure Learning (GSL) focuses on capturing intrinsic dependencies and interactions among nodes in graph-structured data by generating novel graph structures. Graph Neural Networks (GNNs) have emerged as promising GSL solutions, utilizing recursive message passing to encode node-wise inter-dependencies. However, many existing GSL methods heavily depend on explicit graph structural information as supervision signals, leaving them susceptible to challenges such as data noise and sparsity. In this work, we propose \model, an approach that leverages large language models (LLMs) to learn complex node relationships in graph-structured data. By enhancing the reasoning capabilities of LLMs through instruction-tuning over graph structures, we aim to overcome the limitations associated with explicit graph structural information and enhance the reliability of graph structure learning. Our approach not only effectively denoises noisy connections but also identifies node-wise dependencies from a global perspective, providing a comprehensive understanding of the graph structure. We conduct extensive experiments on multiple benchmark datasets to demonstrate the effectiveness and robustness of \model\ across various settings. We have made our model implementation  available at: \url{https://github.com/HKUDS/GraphEdit}.
\end{abstract}

\section{Introduction}
\label{sec:intro}

Graph Structure Learning (GSL) is a burgeoning field of research that strives to unveil the underlying patterns and relationships within graph-structured data~\cite{jin2020graph,fatemi2021slaps}. In GSL, the primary focus lies in unraveling the latent relationships and dependencies that may not be immediately discernible from the raw data. By generating these novel graph structures, GSL empowers us to gain a more comprehensive understanding of the data, thereby facilitating various downstream tasks, such as node classification~\cite{zhao2021heterogeneous}.

In recent years, graph neural networks (GNNs) have indeed captured significant attention and popularity due to their remarkable capacity to model and leverage relationships within graph-structured data~\cite{garg2020generalization,buterez2022graph}. GNNs excel in learning node-level representations by effectively aggregating and propagating information from neighboring nodes in a graph. This exceptional capability has sparked a revolution in the analysis of graph-structured data, enabling a more comprehensive understanding of the underlying node-wise connection patterns and interactions. 

The ability to capture and leverage the intricate dependencies within graphs has undoubtedly propelled graph neural networks (GNNs) to the forefront of graph structure learning~\cite{zhou2023opengsl}. Notably, approaches like SLAPS \cite{fatemi2021slaps}, Nodeformer \cite{wu2022nodeformer}, and GT \cite{shi2020masked} incorporate neural networks that collaborate with GNNs to generate novel graph structures. These models undergo co-optimization, ensuring a seamless and integrated learning process. Moreover, recent studies such as SEGSL~\cite{zou2023se} and CoGSL \cite{liu2022compact} have introduced dynamic methods for learning the graph structure. These approaches adaptively learn the graph structure based on predictions or representations generated by optimized GNNs.

While graph neural networks (GNNs) have demonstrated their high effectiveness, it is important to acknowledge that many of these approaches heavily depend on explicit graph structures, such as node links, as supervision signals for learning accurate representations. However, real-world graph domains often encounter challenges such as data noise and sparsity, which can compromise the reliability of these explicit graph structures. 

To illustrate, let's consider a social network dataset where certain links are missing or incomplete due to privacy settings or limited data availability~\cite{dai2022towards}. Additionally, in recommender systems, the user-item interaction graph may involve cold-start users or items, resulting in highly sparse links~\cite{lin2021task}. Furthermore, various types of bias present in recommender systems introduce noise into the data~\cite{wang2021deconfounded}. In such cases, relying solely on explicit graph structures as supervision signals can lead to representations that are either inaccurate or biased. These challenges necessitate the development of more robust graph structure learning framework that can adapt to and overcome the impact of data imperfections in graph-structured data. \\\vspace{-0.12in}

\noindent \textbf{Contributions}. In light of the challenges outlined earlier, this study seeks to explore how large language models (LLMs) can contribute to reasoning about the underlying graph structures. We introduce our proposed model, \model, which is designed to effectively refine graph structures. Our model's objective is twofold: first, to identify and address noisy connections between irrelevant nodes, and second, to uncover implicit node-wise dependencies. To achieve these goals, our model leverages the rich textual data associated with nodes in graph-structured data. By incorporating the text understanding ability of LLMs, specifically through the instruction-tuning paradigm, we enhance the understanding and representation of graph structures. This allows us to capture implicit dependencies among individual nodes that may not be explicitly encoded in the graph structure itself.

To thoroughly evaluate the performance of \model\ framework, we conducted extensive experiments, comparing it with state-of-the-art solutions. Additionally, we performed an in-depth ablation study and robustness analysis to validate the advantages and rationale behind our model. 


\section{Preliminaries}
\label{sec:model}

\noindent \textbf{Graph-Structured Data}. We define a graph using the tuple $\mathcal{G}=(\mathcal{V}, \mathcal{A}, \mathcal{T})$. Here, $\mathcal{V}$ represents a set of $N=|\mathcal{V}|$ nodes, $\mathcal{A} \in \mathbb{R}^{N \times N}$ is the adjacency matrix that captures the connections between nodes. Additionally, $t_n \in \mathcal{T}$ denotes the textual data associated with each node $n \in \mathcal{V}$ in graph $\mathcal{G}$, which consists of a sequence of $L_n$ language tokens. \\\vspace{-0.12in}

\noindent \textbf{Graph Representation Learning}. focuses on capturing meaningful and informative representations of nodes in a graph, enabling the analysis and modeling of intricate relationships and patterns within the graph data~\cite{buterez2022graph}. In recent years, Graph Neural Networks (GNNs) have emerged as promising approaches for capturing complex node-wise dependencies~\cite{jin2020graph,ji2019graph}. By allowing nodes to exchange information with their neighbors, GNNs update their own representations and facilitate the propagation of information throughout the graph structure, enhancing our ability to understand and analyze the underlying graph data. However, in real-world graphs, noisy and missing connections are prevalent, and they significantly impair the performance of existing graph representation learning methods. \\\vspace{-0.12in}

\noindent \textbf{Problem Statement}. Given the observed graph $\mathcal{G}=(\mathcal{V}, \mathcal{A}, \mathcal{T})$ with noisy structural information, our objective is to improve the graph topology. This involves denoising the noisy connections within the graph data and uncovering the implicit relationships among nodes. By refining the original adjacency matrix $\mathcal{A}$ and obtaining a more informative graph structure $\tilde{\mathcal{A}}$, we can better capture the underlying node-wise dependencies, resulting in an updated graph $\tilde{\mathcal{G}}=(\mathcal{V}, \tilde{\mathcal{A}})$. This refinement process leads to a deeper understanding of the underlying graph structure, thereby improving the performance of downstream tasks by leveraging the updated graph structures $\tilde{\mathcal{G}}=(\mathcal{V}, \tilde{\mathcal{A}})$.

\section{Methodology}
\label{sec:solution}

\begin{figure*}[t]
\centering
\includegraphics[width=1\textwidth]{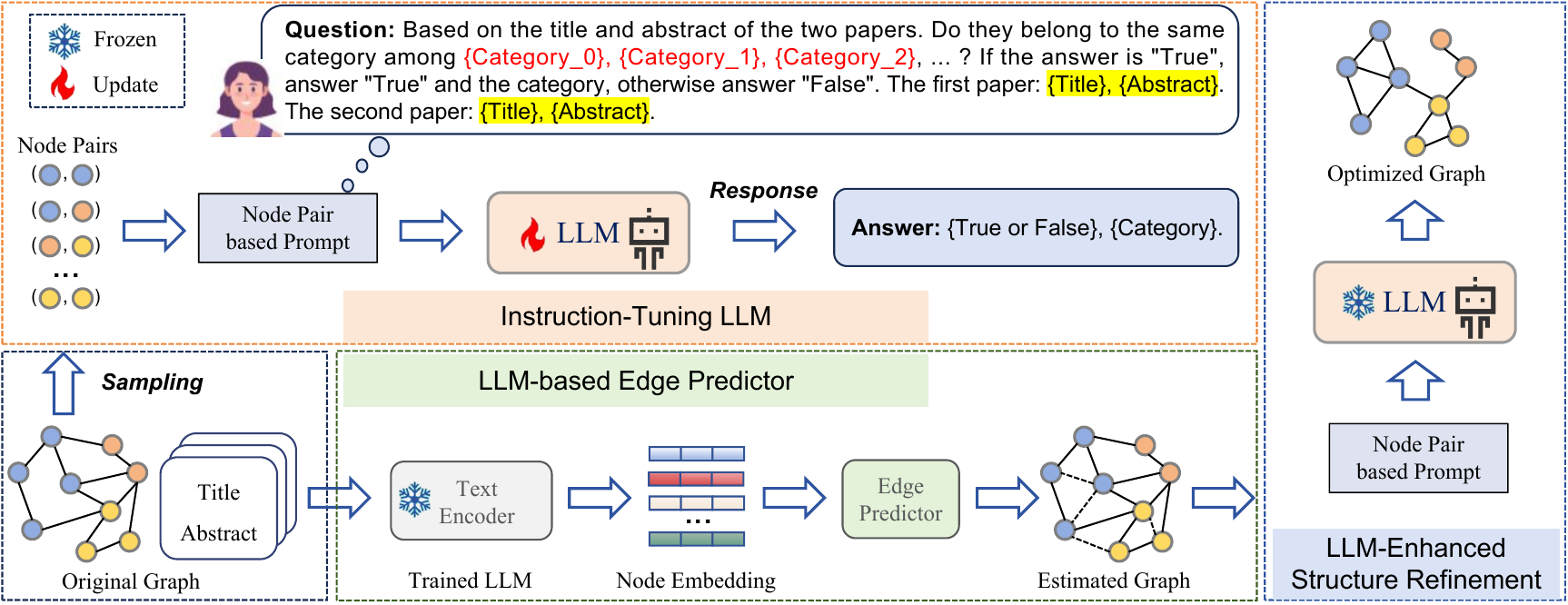}
\vspace{-0.25in}
\caption{The model architecture of our proposed \model\ framework for graph structure learning.}
\vspace{-0.1in}
\label{fig:model}
\end{figure*}

\subsection{Instruction-Tuning LLM}
Taking inspiration from the homophily property assumption discussed in studies such as \cite{gong2023neighborhood,li2023restructuring}, it is suggested that nodes with similar attributes tend to have stronger connections. This concept has further evolved to explore the label consistency between nodes based on their connection patterns \cite{ma2021homophily}. Specifically, in the context of node classification tasks that rely on graph structures, the optimal situation involves maximizing connections within the same class while minimizing inter-class connections. Guided by this principle, our approach aims to leverage the knowledge of Large Language Models (LLMs) to reason about potential dependencies among nodes, taking into account the textual semantics associated with individual nodes.

During the prompt creation phase, we have meticulously defined two separate objectives within each prompt. The first objective is to evaluate the consistency of labels for the node pairs. This objective holds immense importance as it enables the language model to grasp the desired graph structures accurately. The second objective, which builds upon label consistency, involves determining the specific category to which these nodes belong. These carefully crafted prompts, encompassing these dual objectives, serve as valuable resources for instruction tuning of the language model. 

\begin{table}[t]
\centering
\small
\caption{Prompt Instructions for Tuning LLMs.}
\vspace{-0.1in}
\label{tab:ins}
\begin{tabular}{p{7cm}}
\toprule
    Q: Based on the title and abstract of the two paper nodes. Do they belong to the same category among \textbf{\{Category\_0\}}, \textbf{\{Category\_1\}}, \textbf{\{Category\_2\}}, ... ? If the answer is "True", answer "True" and the category, otherwise answer "False". The first paper: \textbf{\{Title\}}, \textbf{\{Abstract\}}. The second paper: \textbf{\{Title\}}, \textbf{\{Abstract\}.} \\
    A: \textbf{\{True or False\}}, \textbf{\{Category\}}. \\
\bottomrule
\end{tabular}
\vspace{-0.15in}
\end{table}

In our methodology, we utilize a random sampling technique to select node pairs $(n_i, n_j)$ from our training data $N_{\text{train}}$. These node pairs, used for tuning the LLMs, are randomly sampled from the training set $N_{\text{train}}$, where $i$ and $j$ represent two distinct nodes, and $k$ represents any other node.
\begin{equation}
\begin{aligned}
    (n_i, n_j) \sim \text{Uniform}(N_{\text{train}} \times N_{\text{train}} - \\
    \{(n_k, n_k) | n_k \in N_{\text{train}}\}).
\end{aligned}
\end{equation}

\subsection{LLM-based Edge Predictor}
To further enhance our analysis, we acknowledge the significance of identifying potential candidate edges in addition to the original graph structure. However, employing the trained language model (LLM) directly to traverse and reason over the entire graph presents a computational challenge, particularly for large graphs, due to the $\mathcal{O}(n^2)$ complexity, $n$ represents the number of nodes in the graph. This computational complexity quickly becomes impractical as the graph size increases. To overcome this challenge, we propose the introduction of a lightweight edge predictor that aids the LLM in the selection process of candidate edges among the nodes in the graph $\mathcal{G}$.

In this approach, we leverage the node pairs that were previously sampled as the training set for the edge predictor. To ensure semantic consistency, we utilize the representations derived from the trained LLM for each node. This can be expressed as:
\begin{equation}
    h_i = \text{LLM}(s_i), h_j = \text{LLM}(s_j),
\end{equation}
We utilize the notation $<i,j>$ to represent a pair of nodes. The textual attributes associated with nodes $i$ and $j$ are denoted by $s_i$ and $s_j$, respectively. The resulting representations, $h_i$ and $h_j$, correspond to their respective nodes and retain the semantic knowledge and reasoning abilities transferred from the large language models.

After obtaining the node representations, we proceed to construct the training set labels $y_e$ based on the node labels $c_n$ using the following procedure:
\begin{equation}
    y_e = 
    \begin{cases}
    1 & \text{if } c_i = c_j \\
    0 & \text{if } c_i \neq c_j
\end{cases}
\end{equation}
Next, we concatenate the representations of the two nodes in each node pair. We then feed the concatenated representation into a prediction layer denoted as $\eta(h_i||h_j)$, which allows us to obtain the probability of the edge's existence. We utilize cross-entropy as the loss function, denoted as $\mathcal{L}_{CE}(y_e,\hat{y}_e)$:
\begin{equation}
    \hat{y}_e = \eta(h_i||h_j) 
\end{equation}
\begin{equation} 
    \begin{aligned}
        \mathcal{L}_{CE}(y_e,\hat{y}_e)&= -[y_elog(\hat{y}_e \\
        &+(1-y_e)log(1-\hat{y}_e)]
    \end{aligned}
\end{equation}
$y_e$ denotes the ground truth label, and $\hat{y}_e$ represents the predicted probability of the edge's existence.

\subsection{LLM-enhanced Structure Refinement}
To refine the graph structure, we employ the previously developed edge predictor to identify the top-k candidate edges for each node based on their estimated likelihood of existence. These candidate edges, along with the original edges of the graph, are then subjected to evaluation by the large language model (LLM) through a prompt, as depicted in Table \ref{tab:ins}. The LLM utilizes this information to determine which edges should be incorporated into the final graph structure. The graph structure refinement process can be summarized as follows:
\begin{align}
    \mathcal{A}' &= \text{EdgePredictor}(\mathcal{H}_n) + \mathcal{A}, \\
    \hat{\mathcal{A}} &= \text{LLM}(\text{Prompt}(\mathcal{A}'))
\end{align}
The updated adjacency matrix, denoted as $\mathcal{A}'$, is obtained by combining the outputs of the edge predictor with the original adjacency matrix $\mathcal{A}$. This fusion process incorporates the edge predictor's predictions into the existing graph structure. Subsequently, the refined adjacency matrix $\hat{\mathcal{A}}$ is generated through the LLM's evaluation of the prompt applied to $\mathcal{A}'$. The LLM leverages its reasoning capabilities to make decisions regarding both the addition and deletion of edges in the final graph structure. Therefore, the refined adjacency matrix $\hat{\mathcal{A}}$ represents the LLM's informed choices, encompassing both the inclusion and exclusion of edges. This refined adjacency matrix serves as an input for downstream graph tasks, e.g., node classification.

In summary, our framework enhances the quality and structure of the final graph by incorporating the edge predictor's predictions and leveraging the reasoning capabilities of the LLM. This leads to the uncovering of implicit global node-wise dependencies and the denoising of noisy connections, resulting in an improved graph representation.
\section{Evaluation}
\label{sec:eval}

\begin{table}[t]
\centering
\caption{Statistics of Experimental Datasets.}
\vspace{-0.15in}
\label{tab:dataset}
\resizebox{7cm}{!}{
\begin{tabular}{ccccc}
\toprule
    & & & & \\[-8pt]
    Dataset & \# Nodes & \# Edges & \# Feat. & Classes \\
    \hline
    & & & & \\[-8pt]
    Cora & 2708 & 5429 & 1433 & 7 \\
    Citeseer & 3186 & 4277 & 3703 & 6 \\
    PubMed & 19717 & 44335 & 500 & 3 \\
\bottomrule
\end{tabular}
}
\vspace{-0.15in}
\end{table}


\subsection{Experimental Settings}
\subsubsection{Datasets} To evaluate the performance of our \model\ method, we carefully selected three representative datasets: Cora, PubMed, and Citeseer. These datasets are widely recognized as benchmarks for graph learning tasks. In these datasets, each node represents a publication, and the edges represent citations between them. i) \textbf{Cora} dataset comprises papers classified into seven computer science domains: Case-Based, Genetic Algorithms, Neural Networks, Probabilistic Methods, Reinforcement Learning, Rule Learning, and Theory. It provides a diverse range of topics within the field of computer science. ii) \textbf{PubMed} dataset focuses on medical literature and categorizes papers into three distinct categories: Diabetes Mellitus, Experimental, Diabetes Mellitus Type 1, and Diabetes Mellitus Type 2. This dataset offers valuable insights into various aspects of diabetes research. iii) \textbf{Citeseer} dataset consists of academic papers from six different areas within computer and information science: Agents, Machine Learning (ML), Information Retrieval (IR), Databases (DB), Human-Computer Interaction (HCI), and Artificial Intelligence (AI).

\subsubsection{Baselines}
To comprehensively validate the effectiveness of our \model\ model, we compare it with 13 graph structure learning baselines, categorized into three groups based on their training strategies.

\noindent \textbf{Pre-Training Models}. GSR \cite{zhao2023gsr}, STABLE \cite{li2022reliable}, and SUBLIME~\cite{liu2022towards} are advanced pre-training models specifically designed to refine graph quality and enhance the effectiveness of graph representation learning. The training process involves two stages: first, enhancing the graph structure through pre-training, and then utilizing this refined structure to train GNNs for various downstream tasks. \\\vspace{-0.12in}

\noindent \textbf{Iter-Training Models}. SEGSL~\cite{zou2023se}, CoGSL \cite{liu2022compact}, and GEN \cite{wang2021graph} employ the iterative training where two components are developed simultaneously. They adaptively learn the graph structure based on predictions or representations generated by an optimized GNN. The learned structure is then used to train a new GNN model in the subsequent iteration. \\\vspace{-0.12in}

\noindent \textbf{Co-Training Models}. Notable examples of co-training models include Nodeformer \cite{wu2022nodeformer}, WSGNN \cite{lao2022variational}, GT \cite{shi2020masked}, SLAPS \cite{fatemi2021slaps}, Gaug \cite{zhao2021data}, IDGL \cite{chen2020iterative}, and GRCN \cite{yu2020graph}. In these models, the neural networks responsible for generating the graph structure are co-optimized alongside GNNs. This co-optimization ensures a more integrated and effective learning process, as both components mutually benefit from each other's improvements.

\subsubsection{Implementation Details}
In our \model\ model, we use Vicuna-v1.5 as our LLM, trained using the LoRA method. The model backbone consists of a two-layer GCN with a hidden size of 128. For our experiments, we divide the Cora, Citeseer, and PubMed datasets into three parts: training, validation, and testing. Following a ratio of 6:2:2, as mentioned in \cite{he2023explanations, tang2023graphgpt, wen2023augmenting}, ensures a consistent approach to dataset division. To train both the LLM and the Edge Predictor, we randomly sample 20,000 node pairs from the training set as training data. During the selection of candidate edges, we experiment with different top-k values ranging from 1 to 5. This exploration enables us to investigate the impact of varying the number of selected edges and determine the optimal setting. To ensure the robustness of our results, we repeat all experiments 10 times and calculate the mean and standard deviation of the outcomes. To facilitate fair comparisons, we tune the parameters of various baselines using a grid search strategy.

\begin{table}[t]
\large
\renewcommand\arraystretch{1.2}  
\centering
\caption{Accuracy comparison between \model\ and various state-of-the-art baselines. These refined structures are then fed into the downstream GCN encoder for representation learning in the node classification task.}
\vspace{-0.1in}
\label{tab:result}
\resizebox{7.5cm}{!}{
\begin{tabular}{cccccl}
\toprule
& & & & \\[-15pt]
\textbf{Model} & \textbf{Cora} & \textbf{Citeseer} & \textbf{PubMed} \\ 
& & & &\\[-15pt]
\hline
GCN & 87.36 $\pm$ 1.60 & 78.87 $\pm$ 2.18 & 87.37 $\pm$ 0.77 \\ 
GRCN & 84.13 $\pm$ 0.37 & 74.23 $\pm$ 1.18 & 85.20 $\pm$ 0.10 \\
IDGL & 88.63 $\pm$ 0.44 & 80.85 $\pm$ 0.07 & 88.30 $\pm$ 0.12 \\ 
GAug & 86.72 $\pm$ 0.63 & 77.61 $\pm$ 1.02 & 84.48 $\pm$ 0.37 \\
GEN & 86.53 $\pm$ 0.63 & 80.38 $\pm$ 0.72 & 87.04 $\pm$ 0.11 \\ 
SLAPS & 81.99 $\pm$ 1.57 & 73.17 $\pm$ 0.87 & 85.21 $\pm$ 0.18 \\
GT & 88.34 $\pm$ 0.35 & 78.46 $\pm$ 0.48 & 86.69 $\pm$ 0.19 \\ 
CoGSL	& 82.07 $\pm$ 0.51 & 78.84 $\pm$ 0.11 & OOM \\
WSGNN	& 89.59 $\pm$ 0.17 & 80.88 $\pm$ 0.48 & 87.17 $\pm$ 0.19 \\ 
SUBLIME & 85.04 $\pm$ 0.37 & 43.73 $\pm$ 7.08 & 86.03 $\pm$ 0.33 \\
STABLE & 88.75 $\pm$ 0.35 & 75.67 $\pm$ 0.98 & 86.30 $\pm$ 0.15 \\ 
Nodeformer & 88.56 $\pm$ 1.01 & 80.28 $\pm$ 0.57 & 87.93 $\pm$ 0.26 \\
GSR & 87.56 $\pm$ 1.19 & 78.77 $\pm$ 1.56 & 85.61 $\pm$ 0.55 \\ 
SEGSL & 87.49 $\pm$ 0.66 & 78.91 $\pm$ 0.52 & 87.57 $\pm$ 0.37 \\
\hline
\model & \textbf{90.90 $\pm$ 1.16} & \textbf{81.85 $\pm$ 1.42} & \textbf{94.09 $\pm$ 0.28} \\
\hline
\end{tabular}
}
\end{table}

\subsection{Performance Comparison}
In our analysis of node classification tasks across three datasets, we compare our \model\ model against various GSL baselines. The results are presented in Table \ref{tab:result}, where "OOM" denotes out of memory error. Following existing GSL methods \cite{zhou2023opengsl, wu2022nodeformer}, we utilize \emph{accuracy} as the evaluation metric. From the comprehensive data analysis, we draw three key observations: \\\vspace{-0.12in}

\noindent \textbf{Obs 1: Remarkable Performance of \model}. Our \model\ model demonstrates superior performance compared to existing graph structure learning methods across the three datasets, establishing itself as a state-of-the-art solution. The remarkable outcomes underscore the capacity of \model\ to enhance graph structures by unveiling implicit global dependencies and efficiently eliminating noisy connections among nodes in a graph. Through this process, \model\ not only improves the accuracy of graph structure learning but also enhances the overall quality and reliability of the learned graph representations. \\\vspace{-0.12in}

\noindent \textbf{Obs 2: Limitation of Existing GSL Approaches}. Among the various GSL baselines, only a subset consistently outperforms the standard GCN, while some even impede the performance of downstream graph representation. These findings shed light on the limitations of alternative solutions that heavily rely on the original graph structures for supervision labels. However, it is crucial to acknowledge that the observed connections between nodes can often be noisy and incomplete, posing challenges for GSL methods in generating high-quality graph representations. In contrast, our \model\ capitalizes on the reasoning capabilities of LLMs to incorporate external semantics into the graph structure learning. By doing so, we enhance the overall quality of the learned representations in downstream tasks with our refined graph structures. \\\vspace{-0.12in}

\noindent \textbf{Obs 3: Performance Variation across Datasets}. When analyzing the performance of \model, we observe a significant improvement on the PubMed dataset compared to Cora and Citeseer. Unlike Cora and Citeseer, PubMed has a larger number of nodes. Consequently, when training with an equal number of node pairs, the LLM encounters a more diverse range of situations in PubMed. Moreover, unlike Cora where missing abstracts are common, the textual information in PubMed's nodes is consistently abundant and detailed. Additionally, with only three categories, the PubMed dataset presents a less complex classification challenge. The same volume of sampling in PubMed enables the LLM to encounter a greater variety of edges associated with each category compared to the other two datasets.

\begin{table*}[t]
\centering
\small
\caption{Model ablation study of our \model\ framework in terms of accuracy.}
\vspace{-0.05in}
\label{tab:ablation}
\resizebox{16cm}{!}{
\begin{tabular}{l|cc|cc|cc}
\toprule
\textbf{Model} & \multicolumn{2}{c|}{\textbf{Cora}} & \multicolumn{2}{c|}{\textbf{Citeseer}} & \multicolumn{2}{c}{\textbf{PubMed}} \\
\midrule
GCN & \multicolumn{2}{c|}{87.36 $\pm$ 1.60} & \multicolumn{2}{c|}{78.87 $\pm$ 2.18} & \multicolumn{2}{c}{87.37 $\pm$ 0.77} \\
MLP & \multicolumn{2}{c|}{77.32 $\pm$ 3.23} & \multicolumn{2}{c|}{71.63 $\pm$ 1.47} & \multicolumn{2}{c}{85.83 $\pm$ 0.62} \\
w/o GNN & \multicolumn{2}{c|}{72.75} & \multicolumn{2}{c|}{68.52} & \multicolumn{2}{c}{90.15} \\
\midrule
Instruction & \textbf{-prompt} & \textbf{-prompt-w/o-ca} & \textbf{-prompt} & \textbf{-prompt-w/o-ca} & \textbf{-prompt} & \textbf{-prompt-w/o-ca} \\
\midrule
\model\ w/o Add & 88.38 $\pm$ 1.06 & 87.90 $\pm$ 1.89 & 80.03 $\pm$ 2.16 & 79.61 $\pm$ 2.20 & 91.83 $\pm$ 0.38 & 90.63 $\pm$ 0.51 \\
\model\ w/o Del & 90.52 $\pm$ 1.26 & \textbf{89.96 $\pm$ 1.25} & 81.43 $\pm$ 1.97 & \textbf{80.19 $\pm$ 2.11} & 89.63 $\pm$ 0.50 & 88.60 $\pm$ 0.60 \\
\model & \textbf{90.90 $\pm$ 1.16} & 89.52 $\pm$ 1.49 & \textbf{81.85 $\pm$ 1.42} & 79.84 $\pm$ 1.95 & \textbf{94.09 $\pm$ 0.28} & \textbf{91.98 $\pm$ 0.45} \\
\bottomrule
\end{tabular}
}
\vspace{-0.1in}
\end{table*}

\subsection{Model Ablation Study}
To analyze the impact of different components on the performance of \model, we conducted ablation experiments from two key perspectives. The experimental results are presented in Table \ref{tab:ablation}. \\\vspace{-0.12in}

\noindent \textbf{Instruction-Tuning Paradigm}. In Table \ref{tab:ablation}, the "-prompt" notation refers to the utilization of a two-stage instruction-tuning paradigm (as shown in Table \ref{tab:ins}) to fine-tune the LLM. These instructions include the task of predicting both the existence of edges and the specific category of connected nodes. On the other hand, "-prompt-w/o-ca" indicates fine-tuning the LLM with simplified instructions that do not involve forecasting the specific node category. \\\vspace{-0.12in}

\noindent \textbf{Graph Structure Refinement}. The "\model\ w/o Add" variant denotes that the \model\ is designed specifically for the deletion of edges from the original graph. This variant's primary objective is to identify and remove unnecessary or irrelevant edges, thus refining the graph structure. In contrast, the "\model\ w/o Del" variant refers to the functionality of the \model\ in appending candidate edges to the original graph structure. This approach aims to enrich the graph by introducing potentially valuable connections between nodes.

Based on the findings presented in Table \ref{tab:ablation}, we can observe three significant phenomena: \\\vspace{-0.15in}

\noindent \textbullet ~(i) These findings emphasize the importance of considering both edge existence and type prediction tasks within the fine-tuning process of \model. The exclusion of node type prediction in the "-prompt-w/o-ca" condition results in a loss of precision in node category matching. \\\vspace{-0.12in}

\noindent \textbullet ~(ii) The performance of "\model\ w/o Add" sheds light on the impact of solely denoising the edges of the original graph, revealing limited effectiveness, particularly in datasets like Cora and Citeseer where the original edges are relatively sparse. However, the significant performance boost observed when adding candidate edges to the GCN highlights the ability of \model\ to capture implicit global dependencies among different nodes. These findings underscore the importance of leveraging both edge deletion and addition strategies, along with the reasoning capabilities of the LLM, to optimize the original graph structures. \\\vspace{-0.12in}

\noindent \textbullet ~(iii) The performance difference between \model\ and "w/o GNN" emphasizes the need to incorporate GNN-encoded structural information for node classification tasks. "w/o GNN" solely relies on LLMs to infer node class based on \model's text understanding ability, without integrating the downstream GNN encoder to preserve graph topology. Therefore, including the GNN encoder is crucial for improved node classification performance as it captures the graph structure within the latent representation space.

\begin{figure}[h]
\centering
\includegraphics[width=0.46\textwidth]{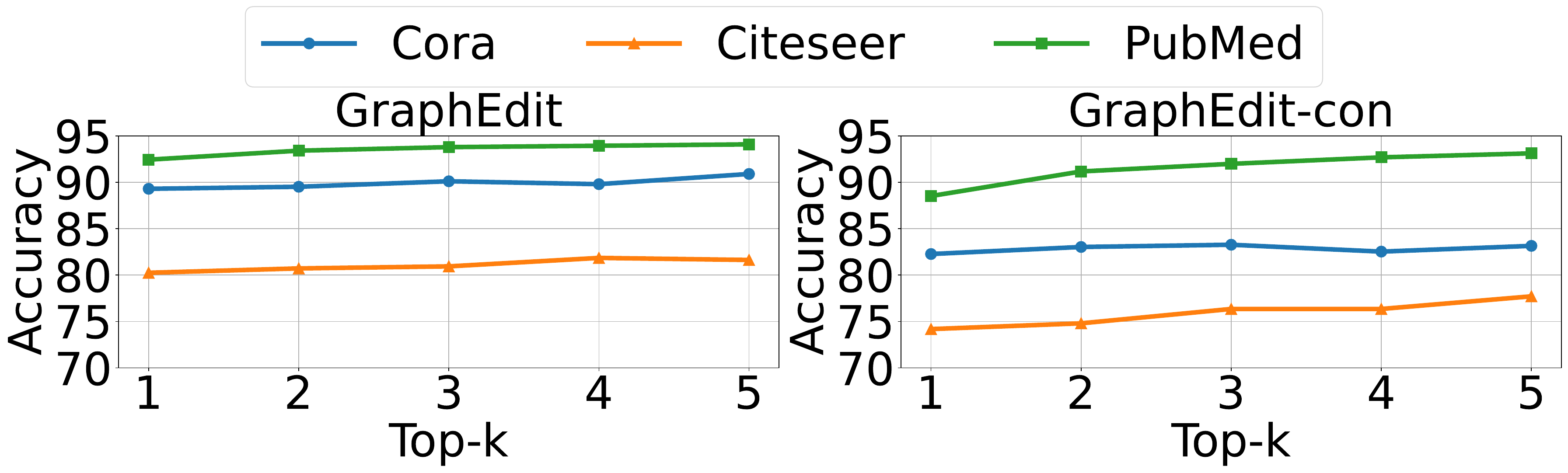}
\vspace{-0.1in}
\caption{Impact study of edge candidate selection.}
\vspace{-0.15in}
\label{fig:topk}
\end{figure}

\subsection{Impact of Edge Candidate Selection}
To investigate the impact of different quantities of candidate edges on the model effectiveness, we analyze the performance across three datasets by varying the $k$-values from 1 to 5 (Figure \ref{fig:topk}). Generally, we observe that higher k-values tend to improve the model's performance. However, on the Cora and Citeseer data, the performance boost plateaus beyond $k=3$, while on PubMed, it stabilizes around $k=4$. This suggests that there is a threshold k-value beyond which the performance of \model\ stabilizes without significant further improvements. These findings provide insights for determining an optimal quantity of candidate edges, ensuring efficient utilization of computational resources while maintaining satisfactory performance.

\begin{table}[h]
\centering
\vspace{-0.05in}
\caption{\model's performance in capturing inherent node relationships without explicit graph structure.}
\vspace{-0.1in}
\resizebox{7.5cm}{!}{
\begin{tabular}{lccc}
\toprule
\textbf{Method} & \textbf{Cora} & \textbf{Citeseer} & \textbf{PubMed} \\ 
\hline
GCN             & \textbf{87.36 ± 1.60}  & \textbf{78.87 ± 2.18}      & 87.37 ± 0.77    \\
MLP             & 77.32 ± 3.23  & 71.63 ± 1.47      & 85.83 ± 0.62    \\
\hline
\model-con   & 83.27 ± 2.02  & 77.71 ± 2.17      & \textbf{93.12 ± 0.29}    \\ 
\bottomrule
\end{tabular}
}
\label{tab:recon}
\vspace{-0.05in}
\end{table}

\subsection{Graph Structure Construction}
To further showcase the capability of our proposed \model\ framework in uncovering implicit node inter-dependencies, we assess its performance on the three datasets without the original graph structure. Results are shown in Table \ref{tab:recon}, where "\model-con" denotes the use of the graph structure constructed by \model\ alone. Remarkably, even in the absence of the original graph structure, \model\ delivers commendable performance. Notably, on the PubMed dataset, \model\ outperforms the original graph structure, highlighting its potential in text-rich scenarios. While \model\ falls short of surpassing the original structure on Citeseer, it achieves comparable results. Thus, this analysis confirms the effectiveness of our model in capturing inherent node relationships, even without an explicit graph structure.

\begin{table}[h]
\centering
\footnotesize
\caption{Performance on the PubMed with injected noisy edges at different rates (0.05, 0.1, 0.15, 0.2, 0.25).}
\vspace{-0.05in}
\label{tab:noise}
\resizebox{7cm}{!}{
\begin{tabular}{ccccccc}
\toprule
\multirow{2}{*}{Method} & \multicolumn{5}{c}{Attack Rate} \\
\cmidrule{2-6}
& 0.05 & 0.1 & 0.15 & 0.2 & 0.25 \\
\midrule
GCN & 86.06 & 85.13 & 84.28 & 83.61 & 83.26 \\
IDGL & 86.20 & 85.29 & 83.54 & 84.18 & 82.39 \\
WSGNN & 85.94 & 85.24 & 84.59 & 83.64 & 84.21 \\
\midrule
\model\ & \textbf{94.07} & \textbf{94.14} & \textbf{94.16} & \textbf{94.20} & \textbf{94.27} \\
\bottomrule
\end{tabular}
}
\vspace{-0.1in}
\end{table}

\subsection{Model Robustness Study against Noise}\vspace{-0.05in}
To investigate the noise resistance of \model, we injected varying proportions of noise (0.05 to 0.25) into the original graph structures of the three datasets. IDGL and WSGNN were selected as benchmarks and subjected to the same noisy conditions. Results are detailed in Table \ref{tab:noise}. The analysis reveals limited noise resistance in IDGL and WSGNN. In contrast, our \model\ method maintains stable performance. Surprisingly, on the PubMed dataset, increasing random noise edges actually improves \model's performance. This suggests effective noise edge elimination while retaining beneficial edges introduced as noise.


\begin{table}
\centering
\small
\caption{Performance comparison with other LLMs.}
\vspace{-0.05in}
\label{tab:LLMs}
\resizebox{7.5cm}{!}{
\begin{tabular}{lcc}
\toprule
\textbf{Model} & \textbf{Cora} & \textbf{Citeseer} \\
\midrule
GCN & 87.36 $\pm$ 1.60 & 78.87 $\pm$ 2.18 \\
ChatGPT 3.5 & 85.30 $\pm$ 2.15 & 78.76 $\pm$ 2.19 \\
ERNIE-Bot-turbo & 86.99 $\pm$ 1.50 & 79.20 $\pm$ 2.25 \\
Vicuna-7B & 87.47 $\pm$ 1.22 & 79.55 $\pm$ 2.17 \\
BLOOMZ-7B & 84.87 $\pm$ 1.58 & 79.47 $\pm$ 2.28 \\
Llama-2-7B & 84.83 $\pm$ 1.94 & 78.65 $\pm$ 1.93 \\
ChatGLM2-6B & 80.92 $\pm$ 2.53 & 74.47 $\pm$ 2.09 \\
AquilaChat-7B & 86.31 $\pm$ 2.05 & 78.17 $\pm$ 2.42 \\
\hline
\model\ & \textbf{88.38 $\pm$ 1.06} & \textbf{80.03 $\pm$ 2.16} \\
\bottomrule
\end{tabular}
}
\vspace{-0.1in}
\end{table}

\subsection{Comparison with other LLMs}
We compared \model\ with commonly used LLMs \cite{brown2020language, sun2020ernie, vicuna2023, yong2022bloom+, touvron2023llama, du2021glm, BAAIAquila2023} to evaluate their denoising capabilities on the original graph structures of the Cora and Citeseer datasets, using the same prompt shown in Table \ref{tab:ins}. The results are summarized in Table \ref{tab:LLMs}. \model\ outperforms other LLMs significantly in denoising on both datasets, demonstrating the effectiveness of our instruction tuning approach. Notably, ERNIE-Bot-turbo, Vicuna-7B, and BLOOMZ-7B perform well on the Citeseer dataset, although their performance is less impressive on the Cora dataset. This discrepancy can be attributed to the frequent occurrence of missing abstracts in Cora's nodes, which hampers the decision-making process of the LLMs and impacts the final graph structure.

\begin{figure}[h]
\centering
\includegraphics[width=0.50\textwidth]{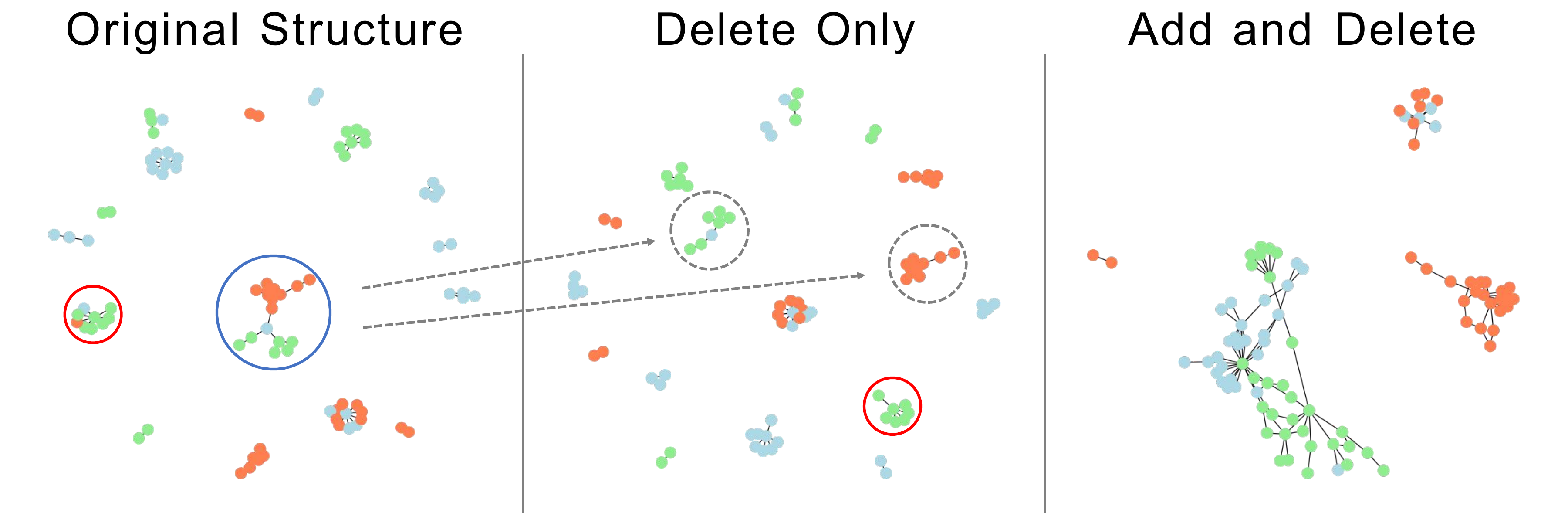}
\vspace{-0.1in}
\caption{Visual analysis with random sampled 20 nodes and their 1-hot neighbors on the PubMed dataset.}
\vspace{-0.15in}
\label{fig:ls}
\end{figure}

\subsection{Visual Analysis}
In this section, we visually compare the original graph structure of PubMed with the optimized graph structure using Figure \ref{fig:ls}. The figures are arranged as follows: the original graph structure is on the left, the \model\ removed structure is in the middle, and the structure after adding and then removing edges is on the right. In the original graph structure, the central node faced classification challenges due to its neighboring nodes belonging to three different categories. However, \model\ effectively addressed this issue by removing the neighbors of different categories around the central node, enabling accurate category determination. 

Additionally, the original structure had a mixed area of three categories, which was successfully split into two distinct substructures after \model\ processing, simplifying the classification task. Moreover, the modified structure on the right maintained intra-class connections while eliminating inter-class links. These observations highlight the ability of \model\ not only to denoise the graph but also to restructure it in a way that greatly facilitates the task of node classification for the GCN.


\begin{table}[t]
\centering
\small
\caption{Case study of \model\ on PubMed data}
\vspace{-0.1in}
\label{tab:case}
\begin{tabular}{p{7cm}}
\toprule
    \textbf{Query:} 
    Based on the title and abstract of the two papers. Do they belong to the same category among Diabetes Mellitus Type 1, Diabetes Mellitus Type 2, or Diabetes Mellitus, Experimental? If the answer is "True", answer "True" and the category, otherwise answer "False". The first paper: Title: \{Node 2601 Title\} Abstract: \{Node 2601 Abstract\}. The second paper: Title: \{Node 6289 Title\} Abstract: \{Node 6289 Abstract\}.  \\
\hline
    \textbf{GT:} 
    True. Diabetes Mellitus Type 1.\\
\hline
    \textbf{The prediction results of our \model:} 
    True. Diabetes Mellitus, Experimental.\\
\bottomrule
\end{tabular}
\vspace{-0.1in}
\end{table}

\subsection{Case Study}
To demonstrate the advantages of predicting node consistency rather than directly predicting node categories, we present a clear example from the PubMed dataset. Table \ref{tab:case} showcases a straightforward case where nodes 2601 and 6289 are connected in the original PubMed graph structure and belong to the same category. During the inference process of \model, although it did not precisely predict the specific category of these two nodes, it successfully identified the consistency of their categories. This instance highlights how the training approach of \model\ effectively reduces the error rate in the LLM's inference, focusing on capturing the underlying consistency rather than precise categorization. This example serves to illustrate the benefits of prioritizing node consistency prediction, emphasizing the ability of the \model\ approach to capture meaningful patterns and relationships in the graph structure, even if it falls short of precisely categorizing individual nodes.

\section{Related Work}
\label{sec:relate}

\noindent \textbf{Graph Structure Learning}.
Various models have been developed to enhance our understanding and optimization of graph structures. Early works like Dropedge \cite{rong2019dropedge} and Neuralsparse \cite{zheng2020robust} focused on graph denoising through edge-dropping. LDS \cite{franceschi2019learning} modeled structures using Bernoulli distributions. More recent approaches, like IDGL \cite{chen2020iterative} and GRCN \cite{yu2020graph}, leverage node representations for structure formation. IDGL employs a weighted cosine function, while GRCN uses dual GNNs for structure derivation. WSGNN \cite{lao2022variational} employs variational inference for joint learning of node labels and graph structure. In contrast, SUBLIME \cite{liu2022towards} explores unsupervised learning with a structure bootstrapping contrastive framework. However, these methods heavily rely on explicit node connections, making them susceptible to data noise and sparsity. \\\vspace{-0.12in}

\noindent \textbf{Large Language Models for Graphs}.
Recent research has begun exploring the application of LLMs in learning with graph-structured data~\cite{wei2023llmrec}. For instance, TAPE~\cite{he2023explanations} utilizes LLMs to predict ranked classification lists for nodes, providing detailed explanations. KEA~\cite{chen2023exploring} enriches node text by incorporating knowledge entities. RLMRec~\cite{ren2023representation} proposes to align GNN embeddings with LLM's knowledge. Additionally, ENG~\cite{yu2023empower} leverages LLMs to generate new nodes, enhancing GNN performance in few-shot learning. GraphGPT~\cite{tang2023graphgpt} investigates the fusion of GNNs and LLMs, developing a customized LLM for graphs. However, none of these approaches address the challenges of noisy and incomplete data commonly found in graphs. In contrast, this work harnesses the reasoning capabilities of LLMs to robustly and effectively optimize the structure of graph-structured data.

\vspace{-0.05in}
\section{Conclusion}
\label{sec:conclusion}
\vspace{-0.05in}
We introduce a groundbreaking large language model called \model, specifically designed for learning graph structures. Our model possesses the remarkable ability to identify noisy connections between nodes and uncover implicit relations among non-connected nodes, thereby enabling the optimization of the graph structure. To achieve this, we seamlessly integrate the power of LLMs with our lightweight edge predictor, which we have developed. This integration empowers our model to refine the graph structures, aligning them with the reasoning knowledge of LLMs. To rigorously evaluate the performance of our model, we conducted extensive experiments across various settings. The results consistently demonstrate the exceptional superiority of \model. Moreover, through thorough investigation, we provide further validation for the rationale behind our model design.

\clearpage

\section{Limitation}
\label{sec:limit}

It is important to acknowledge that, although the results of this study are promising, there are still limitations that need to be addressed in future.

\emph{Firstly}, while our {\model} has demonstrated promising results, it is crucial to examine its performance across a wider range of graph structures. Future studies should explore different types of graph data, such as knowledge graphs and biological networks, to assess the generalizability and adaptability of our model to diverse domains.

\emph{Secondly}, real-world graph structures often undergo changes over time, making it essential to investigate how {\model} can handle dynamic and evolving graphs. Future research should explore strategies to adapt and update the model as new nodes, edges, or attributes are added or modified within the graph. This investigation will enable the model to stay up-to-date and maintain its effectiveness in dynamic environments.

\emph{Lastly}, enhancing the interpretability and explainability of {\model} is an important avenue for further investigation. Developing techniques to extract meaningful insights and explanations from the model's reasoning process will help users better understand and trust the model's decisions.

Addressing these limitations in future studies will not only strengthen the overall understanding and applicability of {\model}. By exploring different graph structures, adapting to dynamic environments, and enhancing interpretability, we can foster the development of more robust and reliable graph-based learning models that can effectively handle a wide range of real-world scenarios.

\bibliography{refs}

\end{document}